\documentclass[10pt,letterpaper]{article}

\usepackage{cogsci}

\cogscifinalcopy
\usepackage[flushleft]{threeparttable}
\usepackage{amsmath}

\usepackage{pslatex}
\usepackage{graphicx}
\usepackage{xcolor}
\usepackage{apacite}
\usepackage{float}

\title{Predicting Human Similarity Judgments Using Large Language Models}

\author{{\large \bf Raja Marjieh$\textsuperscript{1,*}$, Ilia Sucholutsky$\textsuperscript{2,*}$,  Theodore R. Sumers$\textsuperscript{2}$}, \\
{\large \bf Nori Jacoby$\textsuperscript{3}$, Thomas L. Griffiths$\textsuperscript{1,2}$} \\
  $\textsuperscript{1}$Department of Psychology, Princeton University\\
  $\textsuperscript{2}$Department of Computer Science, Princeton University\\
  $\textsuperscript{3}$Computational Auditory Perception Group, Max Planck Institute for Empirical Aesthetics\\
  $\texttt{\{raja.marjieh, is2961, sumers, tomg\}@princeton.edu; nori.jacoby@ae.mpg.de}$ \\
  $\textsuperscript{*}$equal contribution.} 

\begin{document}
\maketitle
\begin{abstract}
Similarity judgments provide a well-established method for accessing mental representations, with applications in psychology, neuroscience and machine learning. However, collecting similarity judgments can be prohibitively expensive for naturalistic datasets as the number of comparisons grows quadratically in the number of stimuli. One way to tackle this problem is to construct approximation procedures that rely on more accessible proxies for predicting similarity. Here we leverage recent advances in language models and online recruitment, proposing an efficient domain-general procedure for predicting human similarity judgments based on text descriptions. Intuitively, similar stimuli are likely to evoke similar descriptions, allowing us to use description similarity to predict pairwise similarity judgments. Crucially, the number of descriptions required grows only linearly with the number of stimuli, drastically reducing the amount of data required. We test this procedure on six datasets of naturalistic images and show that our models outperform previous approaches based on visual information.

\end{abstract}

\section{Introduction}

Mental representations serve as a substrate for a variety of cognitive tasks such as decision-making, communication and memory \cite{anderson2013adaptive}. Understanding the structure of those representation is a core problem in cognitive science and is the subject of a large corpus of work in the psychological literature \cite{shepard1980multidimensional, shepard1987toward, Ghirlanda200315, battleday2020capturing, peterson2018evaluating, jha2020extracting, caplette2022computational, hebart2020revealing}. 

One important example of this research is the development of the multi-dimensional scaling method (MDS) for uncovering the structure of mental representations based on similarity judgments \cite{shepard1980multidimensional}. Given a set of $N$ stimuli, MDS begins by collecting pairwise similarity judgments and aggregating them into a $N \times N$ matrix. Then, an iterative procedure finds an embedding that maps the stimuli into points in a \emph{psychological space} such that their distance mirrors their similarity. Applying MDS to different datasets revealed highly interpretable organization of the stimuli \cite{shepard1980multidimensional, shepard1987toward}. Aside from psychology, similarity judgments play an important role in other disciplines such as neuroscience, e.g., in the method of representational similarity analysis \cite{kriegeskorte2008representational}, as well as in machine learning, e.g., as a way to regularize latent spaces so that they align with human representations and perception \cite{esling2018generative}.

Despite the success of these approaches, the quadratic increase of the number of pairwise comparisons as a function of the number of stimuli poses a serious limitation on their scalability. 
%applicability of methods that rely on similarity matrices to naturalistic domains. 
Indeed, even a relatively small dataset that contains $\sim 10^2$ stimuli would require $\sim 10^4$ judgments for constructing the full similarity matrix. This limitation calls for alternative procedures that allow for efficient approximation of human similarity judgments. Previous studies have proposed such a method in the visual modality by harnessing the latent representations from convolutional neural networks (CNNs) \cite{peterson2018evaluating, jha2020extracting}. Such an approach, however, is domain-specific and could potentially miss important semantic dimensions that weigh on people's judgments. 

To reduce this burden, we leverage the deep relationship between conceptual structure and language~\cite{murphy2002} to use linguistic descriptions as a proxy for human semantic representations.
Intuitively, stimuli that are judged to be highly similar are likely to evoke similar descriptions, allowing us to use description similarity to predict pairwise similarity judgments. This approach offers two key advantages over prior work: first, it is \emph{scalable}. While pairwise similarity comparisons scale quadratically with the number of stimuli~\cite{shepard1980multidimensional}, text descriptions scale linearly. Second, it is \emph{domain-general}: unlike CNN representations~\cite{peterson2018evaluating}, which are limited to visual stimuli, our procedure could be applied to any domain.
%with the number of stimuli which renders them highly scalable ($\sim 100$ descriptions per $\sim 100$ stimuli). 

Finally, we note that our approach leverages two distinct and important advances. First, text descriptions can be easily crowd-sourced via online recruitment platforms such as Amazon Mechanical Turk (AMT; \url{https://www.mturk.com/}) and are part of the common practice in modern machine learning pipelines \cite{parekh2020crisscrossed}. Second, modern language models \cite{speer2017conceptnet, devlin2018bert} provide rich latent representations of text. It is therefore natural to ask: how far can we go in predicting human similarity judgments based on language alone?

We explore this question on a collection of six datasets of naturalistic images for which the ground-truth similarity matrices are known \cite{peterson2018evaluating}. Our exploration proceeds in three stages. In Study 1, we construct similarity estimates by applying a state-of-the-art word embedding model known as ConceptNet NumberBatch (CNNB) \cite{speer2017conceptnet} to pre-existing semantic labels for the dataset images. In Study 2, we generalize this approach by constructing similarity estimates based on BERT, a widely-used large language model \cite{devlin2018bert}, applied to free text descriptions that we crowd-source on AMT. Finally, we combine the concept-level representation of CNNB with the fine-grained textual representation of BERT and generate a joint predictor of similarity judgments. In the process, we benchmark our models' predictive accuracy against the CNN-based approach of \citeA{peterson2018evaluating}.

\section{General Methodology}
\begin{figure}[htp!]
\begin{center}
\includegraphics[width=8.5cm]{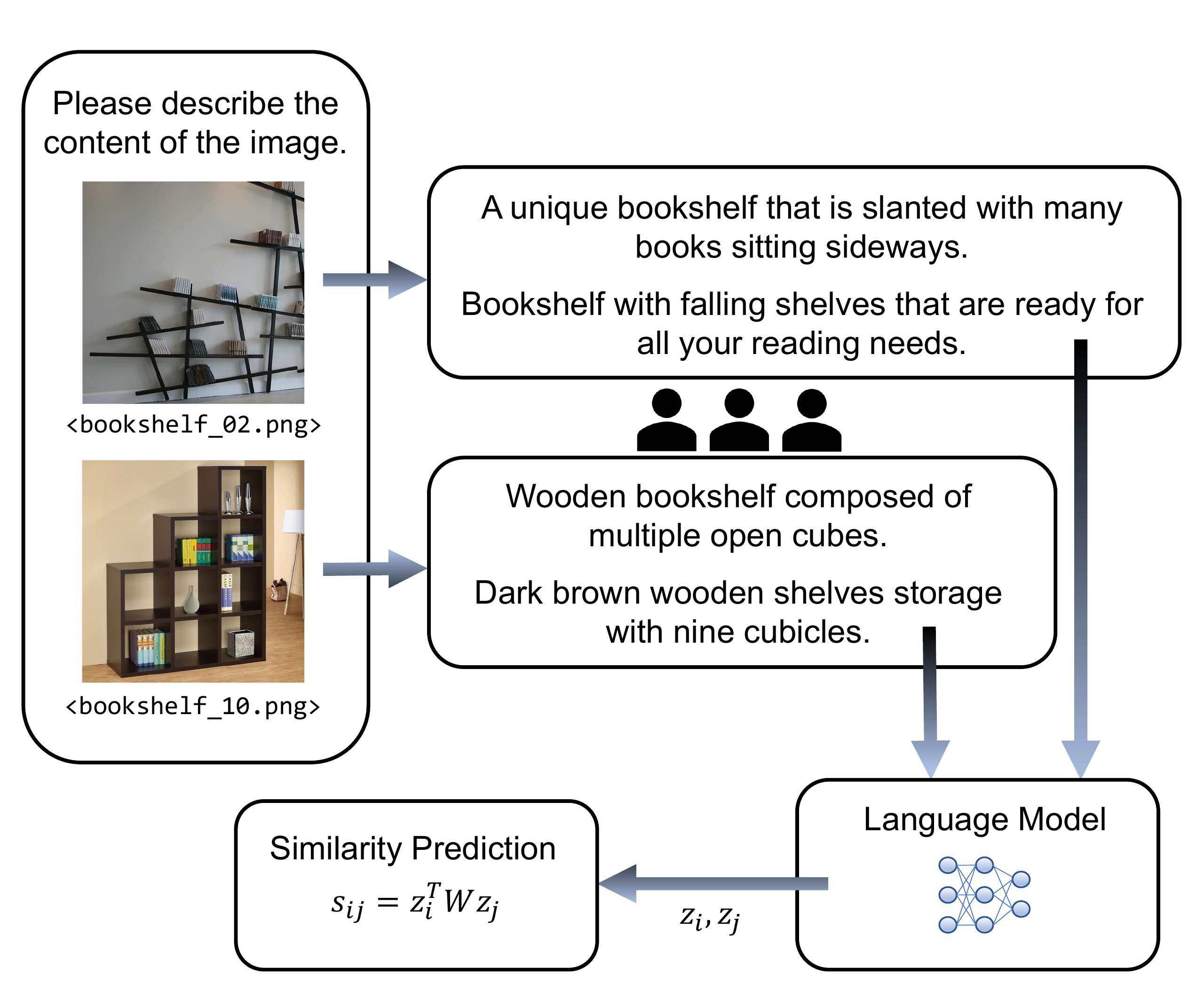}
\end{center}
\caption{Schematic of the similarity prediction procedure based on text descriptions.} 
\label{free-text-schematic}
\end{figure}

Our general pipeline consists of collecting or using pre-existing linguistic descriptors for the individual stimuli and then using an embedding model to compute a proxy for pairwise similarity (Figure~\ref{free-text-schematic}).
\subsection{Predicting Human Similarity}
Given a set of stimuli and their linguistic descriptors (semantic labels or free-text descriptions) as well as a suitable embedding scheme (e.g., a word embedding model) we used cosine similarity between the vectors representing two stimuli as the metric for calculating their similarity (i.e., the dot product of the two embedding vectors divided by the product of their norms). \citeA{peterson2018evaluating} showed that predicting human similarity using CNN representations can be substantially enhanced by linearly transforming those representations. Mathematically, this corresponds to substituting the dot product ${\bf z}_1^T {\bf z}_2$ with ${\bf z}_1^T {\bf W} {\bf z}_2$ where ${\bf W}$ is a suitable diagonal matrix and ${\bf z}_1$ and ${\bf z}_2$ are the embedding vectors. Moreover, Peterson et al.~showed that such a transformation can be found using ridge regression with L2 normalization. We apply this approach to our linguistic representations, using the Python library scikit-learn's RidgeRegression and RidgeCV implementations. To avoid overfitting and simulate generalization in practice, we performed 6-fold cross-validation over images which ensured that no images from the training set are present in the validation set. This ensures that even when combining BERT and CNNB representations, where the number of features increases, overfitting is still avoided. To facilitate comparison with previous work we quantified performance by computing Pearson $R^2$ scores (variance explained) \cite{peterson2018evaluating,jha2020extracting}.

\subsection{Stimuli}
\begin{figure}[htp!]
\begin{center}
\includegraphics[width=8cm]{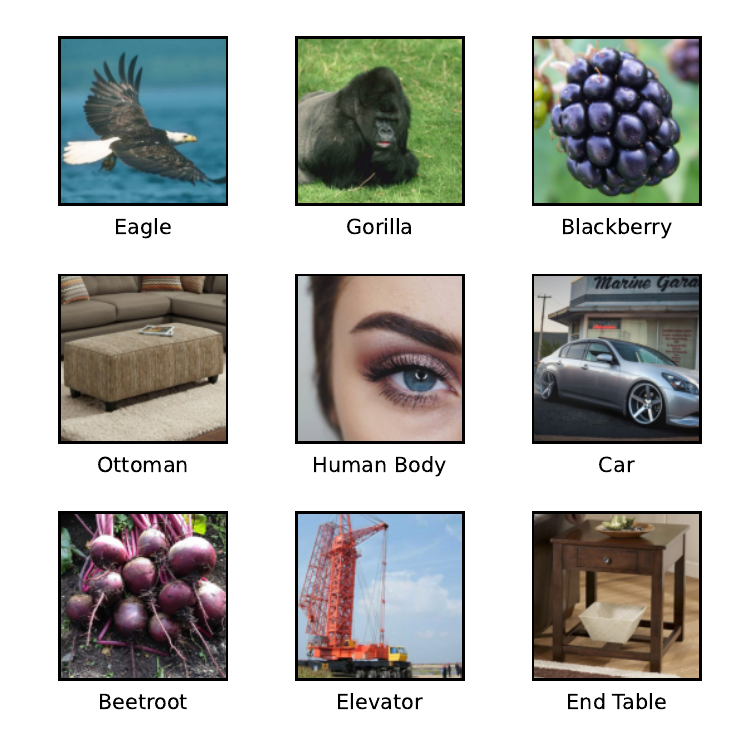}
\end{center}
\caption{Sample images and their semantic labels.} 
\label{sample-images}
\end{figure}

The six image datasets used in this paper were taken from \citeA{peterson2018evaluating}. The datasets were organized based on six broad categories, namely, animals, fruits, vegetables, automobiles, furniture and various objects, each comprising 120 unique images. For all categories except animals, the datasets included semantic labels for each of the individual images. In the case of animals, we manually labeled the images. Sample images and labels appear in Figure \ref{sample-images}.

\section{Predicting Human Similarity \\ Based on Semantic Labels}
To initiate our investigation we first considered using the pre-existing semantic labels for the images in our datasets,  as they served as concise summaries of the content of the images. We evaluated two representations for predicting human similarity judgments based on these labels, namely, a one-hot  representation and a word embedding representation.

\subsection{One-hot Label Representation} 
The first approach served as a baseline and consisted of using the semantic labels as class labels with a ``one-hot" representation, namely, a vector of the form $(0,\dots,0,1,0,\dots,0)$ where the $1$ indicates which semantic label is associated with the image. This representation implies that images with the same semantic label are maximally similar whereas images with different semantic labels are maximally dissimilar.

Surprisingly, this simple representation possessed non-trivial predictive power, as indicated by its average raw $R^2$ score of $0.31$ across the datasets shown in Table~\ref{rsq-table}.

Applying a further linear transformation resulted in a small boost in performance scores ($R^2=0.40$). 
The sparsity of one-hot representations potentially makes linear transformation ineffective. To remedy this, we applied label smoothing to all the one-hot vectors. If $\vec v$ is the one-hot vector, then $\vec v_{smooth}=(1-\epsilon)\vec v + \frac{\epsilon}{k-1}(1-\vec v) $ where $\epsilon$ is the smoothing parameter (we use a value of 0.8) and $k$ is the number of classes (which is equal to the length of the vector). Smoothing does not change the relative structure of the resulting matrix but allows linear transformation to be successfully applied to the new vectors.

Finding positive but not strong correlations is not surprising as the one-hot representation misses fine-grained similarity between related (though not identical) semantic labels. Indeed, although a tiger and a leopard are distinct animals, they nevertheless share some intuitive semantic similarity being members of the cat family; likewise for a chair and a recliner, or a strawberry and a blackberry. This can be seen in the absence of off-diagonal structure in the predicted similarity matrix (Figure~\ref{similarity-matrices}). Nevertheless, this preliminary study serves as an initial evidence for the fact that people's judgments are indeed driven by semantic similarity.

\subsection{Word-embedding Representation}
To capture the structure of similarity between different semantic labels we replaced the one-hot representation with the latent representation of a state-of-art word embedding model known as ConceptNet NumberBatch (CNNB). CNNB is pre-trained on the ConceptNet knowledge graph (\url{https://conceptnet.io/}) which is targeted at capturing intuitive commonsense conceptual relations.

CNNB contains embeddings not only for single words but also concepts consisting of several words. To make use of these, labels consisting of multiple words needed to have spaces replaced by underscores (e.g. `red onion' becomes `red\_onion'). In addition, while the CNNB dictionary is quite large, there are certain words or concepts that it does not contain. In some of these cases, labels consisting of multiple words whose joint form was not found in CNNB had to be separated into individual words and their joint embedding estimated by their normalized sum (e.g. $\text{CNNB(animal body)} \approx \frac{\text{CNNB(animal)+CNNB(body)}}{\sqrt{2}}$). In other cases, labels had to be replaced by a synonym or the closest matching concept available in CNNB (e.g. `tatsoi' was replaced by `spoon\_mustard'). 

\begin{figure*}
\includegraphics[width=\textwidth]{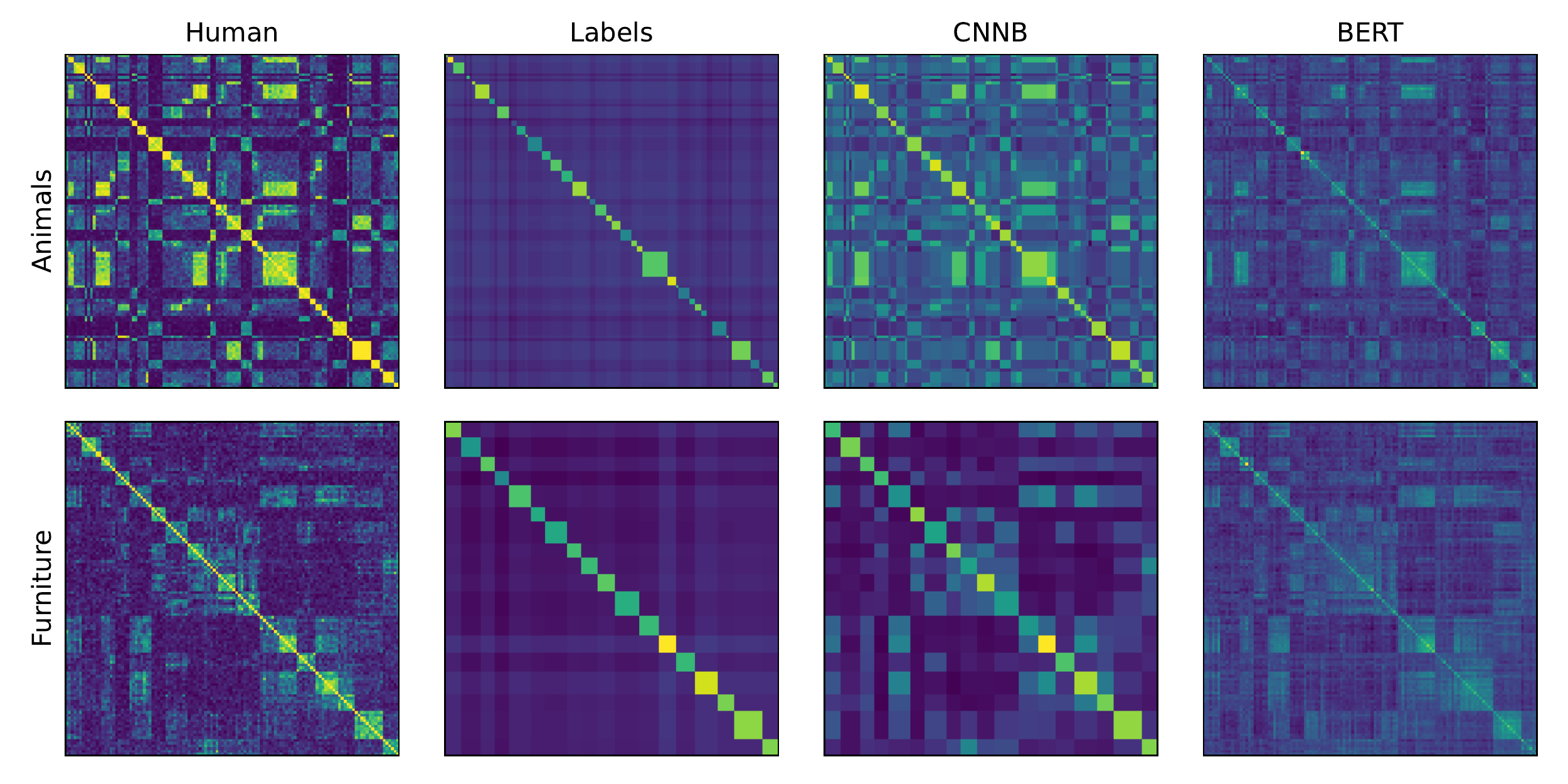}
\caption{Full similarity matrices for the ``animals'' and ``furniture'' datasets for human participants (left), with corresponding predictions based on class labels, CNNB and BERT representations.}
\label{similarity-matrices}
\end{figure*}

The use of CNNB representations resulted in a substantial performance boost over one-hot representations, as reflected in an $R^2$ score of $0.71$ for the transformed representations. The predicted similarity matrix is shown in Figure~\ref{similarity-matrices} and it is clear that a substantial part of the off-diagonal structure is recovered. Similar to the CNN models used by \citeA{peterson2018evaluating}, the linear transformation fine-tunes the broad representations of the model to the specific task at hand. To ensure that the linear transformation is not overfitting the similarity matrices, we performed 6-fold cross-validation as mentioned above and computed a control cross-validated (CCV) $R^2$ score on held-out images. These scores remained high ($R^2 = 0.63$), outperforming the CNN model of \citeA{peterson2018evaluating} (Figure~\ref{model-comp}) on all datasets (except Animals, where it scored lower by a small margin). This implies that CNNB representations generalize better to new data. We also note that the dimensionality of the latent space of CNNB ($d=300$) is much lower than that of the CNN ($d=4096$) reducing the number of possible parameters to optimize over and hence the risk of overfitting.

\section{Predicting Human Similarity \\ Based on Free Text Descriptions}
Concise semantic labels (and corresponding embeddings) are not always available for stimuli of interest.
% describing the content of an image
A more general approach would rely on free-text descriptions, which can be easily crowd-sourced online. Such data, however, requires a different kind of representations capable of flexibly encoding entire sentences (as opposed to aggregating representations of individual words which could lose important within-sentence structure).
To that end, we used the latent representations of BERT~\cite{devlin2018bert}, a popular large-scale language model based on bidirectional transformers, to embed free-text descriptions for each of the individual images which we crowd-sourced on AMT. The data collection procedure as well as example text descriptions are shown in Figure~\ref{free-text-schematic}.  

\subsection{Experimental Methods}

The recruitment and experimental pipeline were automated using PsyNet \cite{HarrisonMarjieh2020}, a framework for experimental design which builds on top of the Dallinger platform (\url{https://github.com/Dallinger/Dallinger}) for recruitment automation. Overall, 328 US participants completed the study and they were paid \$12 per hour. Upon completing a consent form participants had to take a standardized LexTALE English proficiency test \cite{lemhofer2012introducing} to ensure caption quality. Participants that failed to pass the pre-screening test were excluded from the study. Next, participants received the following instructions: ``In this experiment we are studying how people describe images. You will be presented with different images and your task will be to describe their content. In doing so, please keep in mind the following instructions, 1) describe all the important parts of the image, 2) do not start the sentences with ``There is" or ``There are", 3) do not describe unimportant details, 4) you are not allowed to copy and paste descriptions, 5) descriptions should contain at least 5 words, 6) descriptions should contain at least 4 unique words. Note: no prior expertise is required to complete this task, just describe what you intuitively think is important as accurately as possible.'' Participants were then presented with nine random images from the dataset to help give them a sense of the images they were about to describe. 

In each trial of the main experiment participants saw one of the images along with the following prompt ``Please describe the content of the following image" (semantic labels were never provided). They then provided their description in a free text response box, subject to the constraints listed above. Each participant provided up to 30 text descriptions with each image receiving 15 text descriptions on average. To ensure that participants did not provide repetitive responses we computed the average Levenshtein edit distance between their current response and all previous responses. Participants for whom the average distance was close to zero ($<0.2$) after 5 trials were excluded from the study. Any remaining random or very poor quality strings were excluded in a post-processing stage.

\subsection{Computing BERT Embeddings}

We used a pre-trained \texttt{BERT-base-uncased} model with a standard tokenization scheme, accessed via the HuggingFace library~\cite{wolf2020}. For each text description, we first passed the tokens through the BERT model, then took the average embedding across all tokens (e.g. mean-bag-of-words) at each layer. We then averaged the embeddings at each layer across all descriptions for a given image. Empirically, we computed similarity scores based on layers 0 through 12 and picked the best performing layer in each case. In order to combine the BERT and CNNB representations, we first normalized both sets of embeddings by their respective means and standard deviations, and then concatenated the BERT and CNNB embeddings to get a single vector for each image.

\begin{figure}[ht!]
\begin{center}
% \hspace*{-0.22in}
\includegraphics[width=7cm]{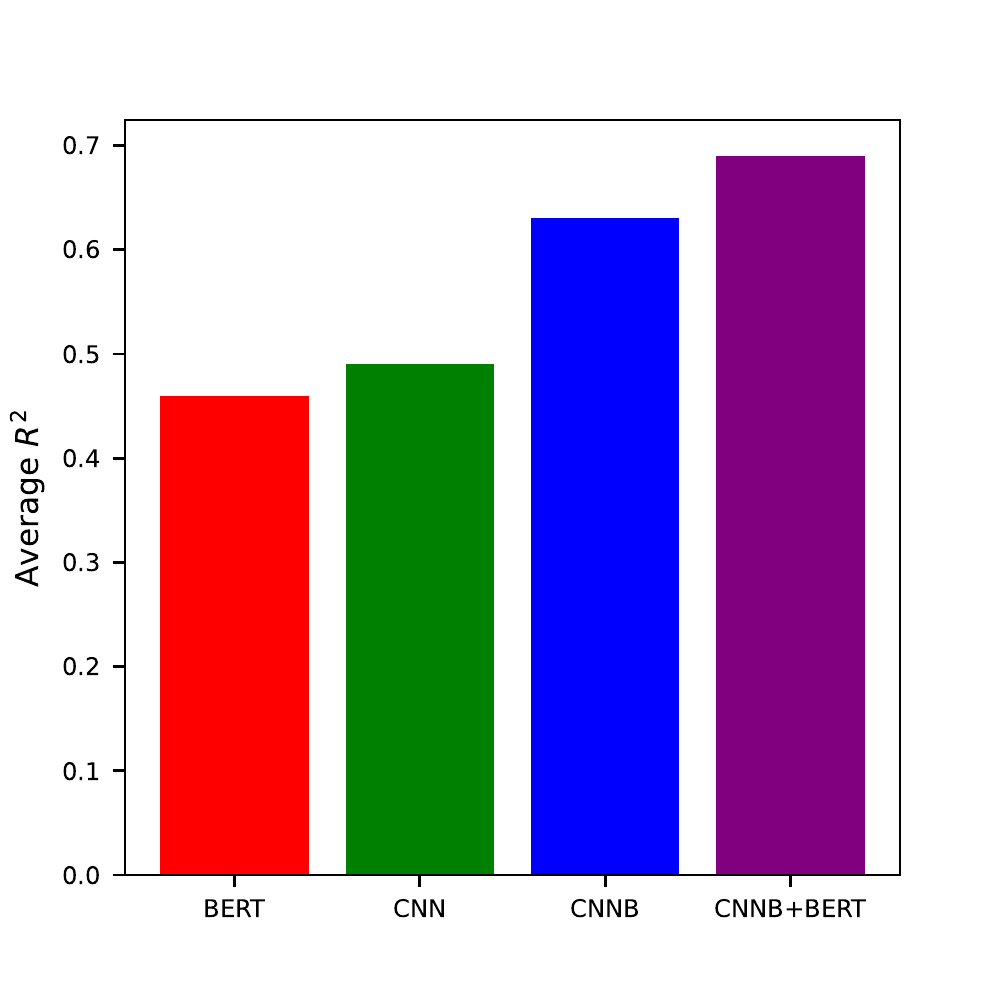}
\end{center}
\caption{Average CCV $R^2$ score for the main four models considered (shown in bold in Table~\ref{rsq-table}).} 
\label{model-comp}
\end{figure}

\begin{figure*}
\begin{center}
\includegraphics[width=16cm]{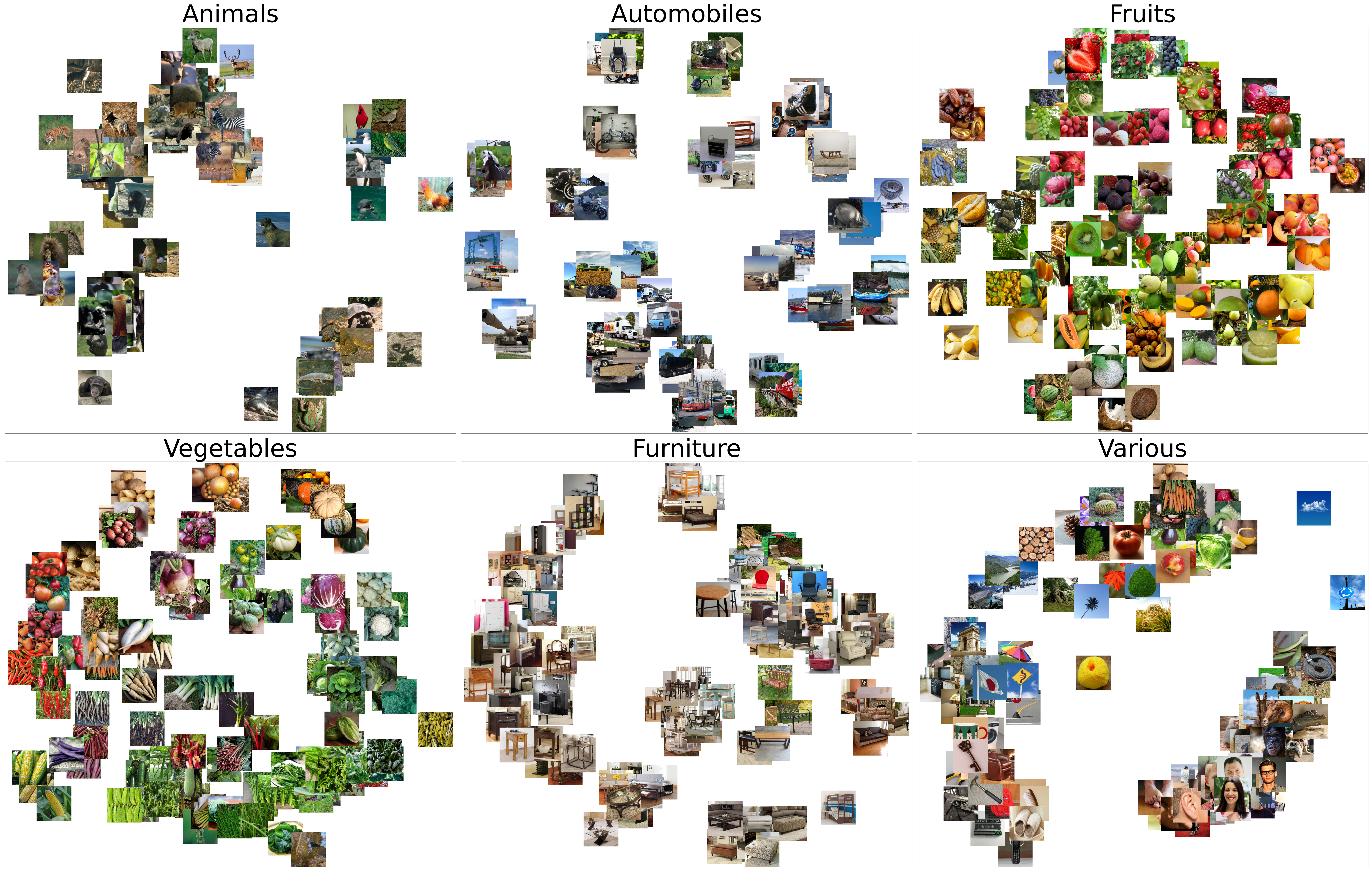}
\end{center}
\caption{Two-dimensional MDS embedding of the joint CNNB-BERT similarity predictions.}
\label{BERT-MDS}
\end{figure*}

\begin{table*}[t]
\begin{center}
\begin{threeparttable}
\caption{$R^2$ scores for the different prediction models and datasets.} 
\label{rsq-table} 
\vskip 0.12in
\begin{tabular}{lcccccccc}
\hline
Model~~~~~    &  Methodology & Animals & Automobiles & Fruits & Vegetables & Furniture & Various & $\langle R^2 \rangle$\\
\hline
Labels~       &   Raw & 0.23 & 0.69 & 0.20 & 0.24 & 0.34 & 0.19 & 0.31 \\
CNNB          &   Raw & 0.51 & 0.64 & 0.17 & 0.17 & 0.31 & 0.29 & 0.35 \\
BERT          &   Raw & 0.22 & 0.30 & 0.09 & 0.13 & 0.25 & 0.36 & 0.23 \\
CNN*          &   Raw & 0.58 & 0.51 & 0.27 & 0.19 & 0.37 & 0.27 & 0.37 \\
Labels~~~~~   &   LT-Train & 0.29 & 0.71 & 0.26 & 0.27 & 0.38 & 0.48 & 0.40\\
CNNB~~~~~     &   LT-Train & 0.85 & 0.86 & 0.53 & 0.60 & 0.67 & 0.72 & 0.71 \\
BERT~~~~~     &   LT-Train & 0.79 & 0.75 & 0.55 & 0.64 & 0.61 & 0.80 & 0.69 \\
CNN*~~~~~     &   LT-Train & 0.84 & 0.79 & 0.53 & 0.67 & 0.72 & 0.52 & 0.68 \\
\textbf{CNNB}     &   \textbf{LT-CCV} & \textbf{0.72} & \textbf{0.86} & \textbf{0.38} & \textbf{0.43} & \textbf{0.63} & \textbf{0.73} & \textbf{0.63} \\
\textbf{BERT}~~~~~     &   \textbf{LT-CCV} & \textbf{0.52} & \textbf{0.53} & \textbf{0.23} & \textbf{0.40} & \textbf{0.47} & \textbf{0.62} & \textbf{0.46} \\
\textbf{CNNB + BERT} &   \textbf{LT-CCV} & \textbf{0.74} & \textbf{0.85} & \textbf{0.44} & \textbf{0.54} & \textbf{0.64} & \textbf{0.76} & \textbf{0.66} \\
\textbf{CNN}*~~~~~      &   \textbf{LT-CCV} & \textbf{0.74} & \textbf{0.58} & \textbf{0.36} & \textbf{0.35} & \textbf{0.35} & \textbf{0.54} & \textbf{0.49} \\
\hline
\end{tabular}
\begin{tablenotes}
  \item Note: ``Raw" corresponds to raw representations, ``LT-Train" corresponds to linearly transformed representations evaluated on training set, and ``LT-CCV" corresponds to linearly transformed representations evaluated on held-out images. $\langle R^2 \rangle$ is the average $R^2$ across all datasets. * indicates values reproduced from \citeA{peterson2018evaluating}.
\end{tablenotes}
\end{threeparttable}
\end{center} 
\end{table*}

\subsection{Results}

We used the embeddings to produce similarity estimates as before. We found that while the raw representations of BERT did not constitute a strong predictor, the linearly re-weighted BERT representations ($d=768$) demonstrated generalization performance comparable to the CNN-based model ($d=4096$) of \citeA{peterson2018evaluating} (Figure~\ref{model-comp}), though not as high as CNNB. One possible explanation for this difference is that CNNB predictors used single concise labels per image whereas for BERT we averaged representations of multiple descriptions which could capture different aspects of the image~\cite{parekh2020crisscrossed}. A more sophisticated approach could learn to pool embeddings from different descriptions efficiently; however for the purpose of the current work we chose to focus on simple linear transformations.

As a last step, we constructed a combined predictor that stacked CNNB and BERT representations to capture broad concept-level knowledge as well as fine-grained descriptions. The combined model resulted in the best aggregated performance, improving further on the CNNB model (Figure~\ref{model-comp}). 

To appreciate the semantic content of the predicted similarity matrices, we computed a two-dimensional MDS representation of the images.  These representations were computed using the scikit-learn library with a maximum iteration limit of $10,000$ and a convergence tolerance of 1e-100. First metric MDS was applied to get an initial embedding, then four iterations of non-metric MDS were applied and the best solution was picked.
The results are shown in Figure~\ref{BERT-MDS}, and reveal a rich and interpretable semantic organization of the stimuli capturing a variety of semantic dimensions such as natural and functional classes as well as color gradients.

\section{Discussion}
 We proposed a highly efficient and domain-general procedure for predicting human similarity judgments based on text descriptions with linear (as opposed to quadratic) complexity. We tested our approach on six datasets of naturalistic images, finding evidence for its validity as well as outperforming previous models that rely on CNNs. These results suggest that human similarity judgments are indeed grounded in semantic understanding and language. Beyond the immediate potential for scaling up studies of similarity, our work also provides a new perspective on the representational similarity between BERT and humans~\cite{lake2021word}: when tested on naturalistic datasets with freely generated text descriptions, we find that BERT successfully captures a substantial part of the structure of human similarity judgments.
 
\begin{figure}[ht!]
\begin{center}
\hspace*{-0.22in}
\includegraphics[width=8cm]{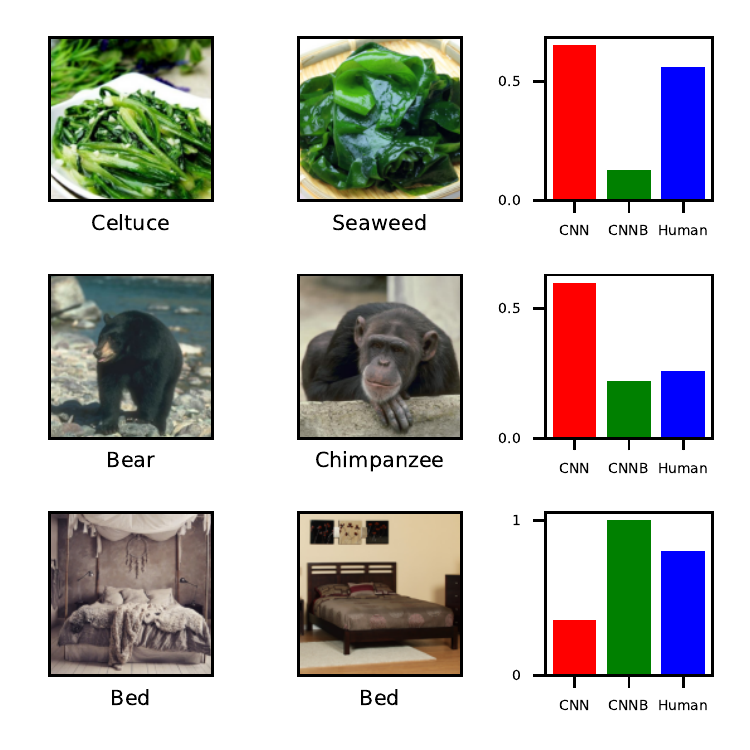}
\end{center}
\caption{Examples of image pairs that generated large discrepancies between CNN and CNNB model predictions and their relation to human similarity scores.} 
\label{disc-images}
\end{figure}

 This work represents an initial step towards a broader investigation of similarity in naturalistic domains. First, our approach offers the possibility of predicting human similarity in other domains such as audio and video. Second, it could be used to explore differences between perceptual similarity (based on raw judgments) and semantic similarity (based on text descriptions). This discrepancy may vary by domain or expertise. For example, in the musical domain, experts (e.g., trained musicians) may provide rich descriptions of stimuli (e.g., musical chords) while novices may lack an appropriate vocabulary, yielding a bigger gap between perception and semantics for the second group. A fine-grained study of this gap as a function of expertise could be informative about the trajectories of semantic development. Third, a systematic study could, for example, use CNN and CNNB representations as a way of isolating perceptual and semantic contributions to a human similarity judgment. Of particular interest are cases of maximal discrepancy whereby humans align with one of the predictors but not the other. Figure~\ref{disc-images} shows examples of such pairs. These seem to suggest that people tend to focus on low-level perceptual features when the objects of comparison are unfamiliar, whereas they would neglect these for familiar objects. A future study could explore this hypothesis in greater detail.
 
In addition to psychological applications, our paradigm may allow for advances in machine learning. Enriching machine learning datasets with similarity judgments and behavioral data more generally can endow artificial models with a variety of useful properties, such as robustness against adversarial attacks and human alignment \cite{peterson2019human}. Collecting similarity judgments over all pairs is infeasible for such datasets due to the large number of stimuli. Nevertheless, in many real-life applications similarity matrices tend to be sparse, i.e., only a small subset of comparisons would yield non-vanishing similarity \cite{parekh2020crisscrossed}.
An efficient enrichment pipeline, therefore, must exploit this sparsity and our procedure is a promising candidate for guiding such methods by predicting which pairs are likely to be informative \emph{a priori}. 
Second, for more domain-specific applications, a followup study could leverage recent advances in multi-modal transformer representations to construct better similarity metrics by incorporating both visual and semantic cues. 
We hope to engage with all of these avenues in future research.

\vspace{2mm}

\noindent {\bf Acknowledgments.} This work was supported by a grant from the John Templeton Foundation.

\bibliographystyle{apacite}

\setlength{\bibleftmargin}{.125in}
\setlength{\bibindent}{-\bibleftmargin}

\bibliography{main}

\end{document}